# PrecTime: A Deep Learning Architecture for Precise Time Series Segmentation in Industrial Manufacturing Operations

1 Stefan Gaugel (Department of Factory of the Future, Bosch Rexroth AG, Ulm, Germany)
Contact: stefan.gaugel@boschrexroth.de

2 Manfred Reichert (Institute of Databases and Information Systems, University of Ulm, Germany)

*Abstract*— **The fourth industrial revolution creates ubiquitous sensor data in production plants. To generate maximum value out of these data, reliable and precise time series-based machine learning methods like temporal neural networks are needed. This paper proposes a novel sequence-to-sequence deep learning architecture for time series segmentation called PrecTime which tries to combine the concepts and advantages of sliding window and dense labeling approaches. The general-purpose architecture is evaluated on a real-world industry dataset containing the End-of-Line testing sensor data of hydraulic pumps. We are able to show that PrecTime outperforms five implemented state-of-the-art baseline networks based on multiple metrics. The achieved segmentation accuracy of around 96% shows that PrecTime can achieve results close to human intelligence in operational state segmentation within a testing cycle.**

*Index Terms*— **Artificial Intelligence, Big Data, Deep Learning, End-of-Line-Testing, Hydraulics, Machine Learning, Neural Network, Time Series Segmentation**

## 1. Introduction

### 1.1 Problem Statement

The past few years have seen a digital transformation that has generated many technological opportunities, one of the most prominent being the widespread use of Artificial Intelligence (AI) and big data-driven solutions in manufacturing. Especially in cyber-physical manufacturing systems, where a large amount of structured data is collected and stored daily, the use of AI methods such as deep learning can be highly beneficial. This is referred to in scientific literature as 'smart manufacturing', which refers to the combined use of various technologies, such as Machine Learning, Industrial Internet of Things (IIoT), Digital Twins and Cloud Computing, for manufacturing and logistics in factories or production sites [1]. In today's global and highly competitive markets, companies need to run their production as cost-effectively as possible, especially when producing in high-wage industrialized countries. This cost pressure is driving the fourth digital revolution to gain momentum rapidly. The amount of sensor data collected in factories is increasing rapidly, and thus, the need to find efficient methods for utilizing this data is becoming increasingly urgent. Historically, the fields of image recognition and natural language processing have been the main focus and forerunners of deep learning, whereas time series-based AI methods are still lagging behind, despite ongoing progress in recent years. One of the biggest challenges remains time series segmentation, especially if it is defined as automatically assigning labels to different segments of a time series and is not limited to merely finding temporal breaking points in a sequence. As the term is not used consistently in literature, this article defines time series segmentation as the complete process of splitting up the time series into an unknown number of non-overlapping segments and automatically labeling them.

Even though time series segmentation has received considerably less attention compared to other time series-related tasks such as classification or forecasting, its importance should not be underestimated. This is especially true in the context of AI-based process automation and quality control, which are major goals of the fourth industrial revolution. In a production setting, one main application for time series segmentation is Operational State Detection (OSD) in multi-phased machining or testing processes. Typically, Machine Learning methods such as statistical-based feature extraction or anomaly detection do not perform well when applied directly to multi-phased data. This is due to the fact that the underlying characteristics of the process data vary greatly between phases, and each measured value must be evaluated in the context of the associated process phase. Therefore, time series segmentation must be performed in advance to structure the process data and enable value-generating follow-up algorithms to succeed. In machinal OSD, these follow-up tasks may include state-specific machine health evaluations or process anomaly detection, as well as the implementation of differentiated storage strategies for data collected in different phases of a manufacturing process. Another application of time series segmentation is Human Action Recognition (HAR), which focuses on understanding the behavior of human operators and gaining information for process optimization tasks.

Many recently proposed end-to-end approaches for time series segmentation are based on deep learning architectures. While deep learning has not established its dominance in time series segmentation as clearly as in other fields such as natural language processing, the state-of-the-art has been considerably pushed forward by neural network architectures in recent years. In particular, the fields of sleep staging and HAR have achieved a pioneering role in this area [2,3]. However, some problems remain unresolved. In most works, the segmentation approach requires the time series to be split into a high number of evenly-sized time window samples and then to assign a label to each of these samples before re-merging them. While these so-called sliding-window approaches lead to satisfactory results in many use cases, especially when the label-change frequency is low, they come with the problem that the granularity of the label predictions is limited by the time window size and stride. Exact time series segmentation approaches that do not make use of a sliding window transformation are often referred to as dense labeling and are less frequently found in literature. Usually, they are based on fully convolutional architectures, sometimes supported by occasional recurrent layers. These dense labeling approaches operate at the highest granularity level by providing each time step with a distinct class prediction (see Figure 1).

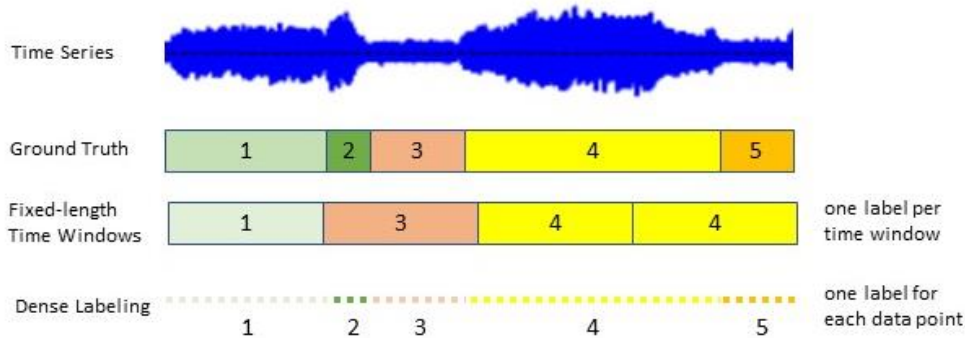

Fig. 1. Sliding window labeling vs dense labeling

### 1.2 Our Contribution

Existing dense labeling approaches come with several drawbacks like struggling to incorporate long-term contextual information when deciding for a label, low training speed, and lack of robustness. However, in this article we analyze an OSD-problem where it is necessary to include the context of the sensor data while at the same time making predictions at a very high granularity. Therefore -- to overcome the shortcomings of the existing approaches -- this article proposes a novel deep learning sequence-to-sequence architecture for precise time series segmentation mixing sliding window and dense labeling concepts to combine the advantages of both:

- The architecture builds upon the state-of-the-art sliding window framework found in the sleep staging domain [4], expanding the two-module framework which consists of an intra-window feature extractor and a Long Short-Term Memory (LSTM)-based inter-window context detector by a third module. The added module is a Convolutional Neural Network (CNN)-based and used for prediction refinement to create dense labels instead of one label per time-window, thereby combining the benefits of sliding-window approaches (including context information, robustness) with the high granularity of dense labeling methods.
- The network uses the concept of multistage labeling by including two loss layers. The window-wise low-granularity prediction of the context detection module is considered as a first stage prediction and connected to a first loss layer. The second loss layer is added after the final prediction output of the model. The two loss layers create a stable learning process for the whole network and enforce it to work according to the idea that the third module just refines the already mostly correct prediction of the first stage

The proposed architecture is a general-purpose framework for time series segmentation that can be used for various industrial applications. In this article, we evaluate the architecture in the experimental section using a hydraulic pump End-of-Line (EoL) testing dataset. In EoL testing, a product is forced through a testing cycle at a testbench to determine its functional quality before its delivery. The testing cycle of a pump consists of various, often fast-changing operational states which have to be precisely distinguished. A precise state distinction is necessary for automating the quality assessment process of the pumps. Without reliable and precise segmentation, necessary follow-up tasks like the intelligent evaluation of collected vibration data and characteristic curves as well as automated tolerance comparison will be erroneous and unstable. A more detailed description of the data and the use case can be found in Section 4. Altogether, the contributions of this paper are twofold:

1. A new general-purpose architecture for precise time series segmentation, called PrecTime, is presented. PrecTime combines the advantages of time-window labeling with dense labeling (main contribution)
2. The approach is applied and validated on a new dataset in the field of hydraulic End-of-Line testing to solve a real-world industrial use case. State detection in EoL testing cycles represents a novel application for deep learning in manufacturing.

The remainder of this paper is structured as follows: Section 2 presents the results of our literature study, Section 3 introduces the PrecTime architecture in detail, Section 4 describes the hydraulic pump use case and the dataset used to evaluate the PrecTime architecture. Section 5 discusses the results, limitations, and future research directions. Finally, Section 6 concludes and summarizes the research.

## 2 Literature Review

The literature review is divided into two subsections, namely time series segmentation research (methodical side) and operational state detection research (application side).

### 2.1 Time Series Segmentation

As the concepts of the IIoT have been increasingly implemented in practice, the amount of available time series data gathered through sensors is growing as well. In turn, this is boosting the research on multivariate time series data as well as the methods to process and extract information from them, including time series segmentation. Before however, the field of time series segmentation has also been thoroughly studied in other domains, see e.g. economics and stock market analysis [5], sleep staging [2], human action recognition [3], speaker diarisation [6], and climate studies [7]. Note that most works covering time series segmentation have a clear domain-specific focus, although emphasizing that their methods should be generally applicable. Only a limited number of works tackle the general problem of time series segmentation. The main forerunners in the last years have been the fields of sleep staging and HAR, both providing general segmentation frameworks with good experimental results and interesting insights.

During the last decades, many different approaches have been proposed, including Hidden Markov Models [8], dynamic time warping [9], and classic feature extraction ML-learning algorithms [10]. However, only with the advent of efficient neural network architectures (CNN, LSTM) the field could make significant progress in the last years utilizing deep learning-based approaches. Thus, we focus on deep-learning-based time series segmentation in the following.

As the terminology and understanding of time series segmentation is not uniform across various fields, two different forms are distinguished: The first one is often referred to as changepoint (or breakpoint) detection and aims to identify structural breaks in time-series (e.g, [11, 12]) . The second form is denoted as segment labeling by us and advances changepoint detection by targeting to directly provide a segment label for each data point in the time series (see Figure 2). As this paper presents a segment labeling use-case (Section 4), solely previous time series research on segmentation labeling is considered in the literature review.

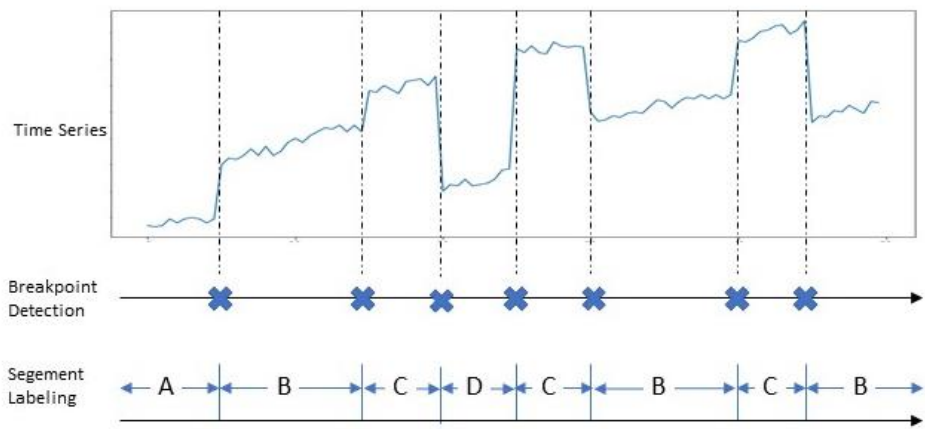

Fig. 2. Time series segmentation: breakpoint detection and segement labeling

There are two main approaches for segment labeling: sliding window-based labeling (alternatively called: epoch-wise labeling) and dense labeling (alternatively: fine-grained recognition, precise segmentation), with the former being much more widespread. Transforming the original time series into equally-sized windows before labeling them simplifies the segmentation process and leads to accurate and stable results, if the window size can be chosen small enough to achieve sufficient granularity. A fundamental distinction here has to be made between classic one-to-one and more advanced sequence-to-sequence methods. One-to-one models only take one time window at a time as input and provide one labeling prediction for this time window. Sequence-to-sequence networks, in turn, transform one sequence into another. In the context of sliding window-based approaches this means that several time windows are fed to the model simultaneously and, thus, the output consists of a sequence of predictions as well, usually one label for each time window given to the network (see Figure 3).

One-to-one architectures constituted the state-of-the-art until around 2017. Essentially, they transform time series segmentation into a series of distinct time series classification steps. [13] presented an LSTM-CNN-based network for human activity recognition based on multivariate time series data. [14] introduced deep learning into sleep staging and evaluated different CNN-based one-to-one architectures for automatic sleep stage labeling. Another work proposed the usage of an attention-based network to extract features and classify a sleep stage epoch more accurately [15]. In the field of speaker diarisation, an LSTM-based network was shown by [16] to successfully classify time windows before remerging them again to achieve the prediction for the whole sequence.

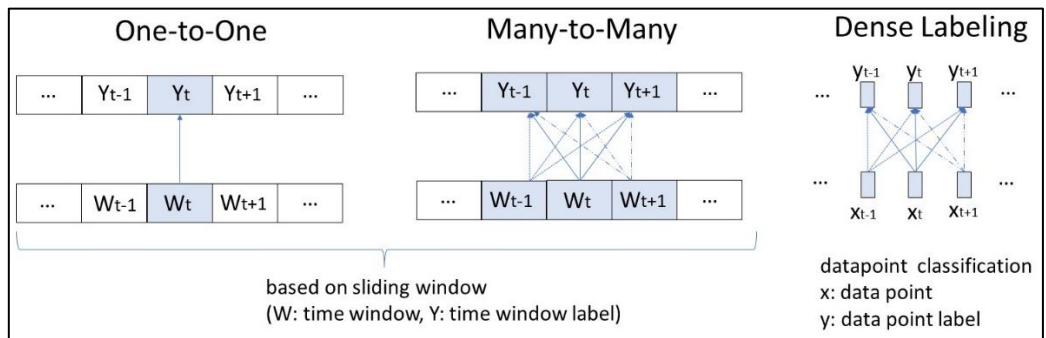

Fig. 3. One-to-One, Many-to-Many, and Dense Labeling Approaches

In recent years, sequence-to-sequence (also called many-to-many) architectures outperformed one-to-one models. This was explained by the fact that sequence-to-sequence architectures are able to capture long- and mid-term contextual information in the time series and are not limited to the information found within one time window. Especially in the field of sleep staging, much progress was made concerning sequence-to-sequence architectures. One of the first works dealing with this topic proposed an architecture, called DeepSleepNet, which uses CNN-layers to extract features inside a time-window and two bi-directional LSTM-layers to capture the contextual inter-widow information of a sequence [17]. Two seminal works were published later by Phan et al. [4, 18] who presented more generic sequence-to-sequence deep learning frameworks based on an epoch-wise feature extraction module and RNN-based inter-epoch processing module. An advancement of their architecture was proposed in 2022 when it was suggested to extend the framework by including transformer modules to achieve more efficient feature extraction and context detection [19]. These approaches achieved high scores in sleep staging, where labels change only once-in-a-while and the different segments are usually far longer than a time-window. However, in other fields very frequent label changes are common and segments may have a very short duration leading to the problem that not all data points within a time window may share a common label. In these cases, a very high granularity of label prediction becomes necessary which dense labeling approaches can provide.

Dense labeling approaches are less frequently found compared to sliding window approaches; they achieved their highest attention in the field of Human Action Recognition. By definition, all dense labeling applications are sequence-to-sequence approaches, since their core principle is to classify each data point of a whole sequence separately without using a sliding window concept. The most relevant contributions include [20], which proposes a simple dense-labeling architecture based on CNN-layers and [21], which presents a multi-stage time convolutional network inspired by already existing video segmentation approaches. Another architecture is proposed [22], which relies on the combination of CNN- and RNN-layers to include temporal as well as spatial information and evaluate their model on a HAR dataset. However, dense labeling approaches are not limited to HAR; an important work in the sleep staging field was published in 2015 [3], which adapted the U-net architecture to the time series domain and created U-time, a generally applicably time series framework, and applied it to sleep staging.

As can be seen, sliding window and dense labeling approaches have mostly been two distinct, non-overlapping fields with different ideas and applications. In this paper, we aim to overcome this distinction and connect the approaches from both fields. To the best of our knowledge, no previous work has used the general architecture of sliding window architectures and extended it to enable dense label predictions. Therefore, we aim to combine the strengths of the two concepts.

## 2.2 Automated Operational State Detection

Since EoL-testing cycle segmentation is a form of OSD, we also had a look on past research about OSD. Automated, sensor-based OSD is considered as an important task in many different settings and industrial applications. Existing works cover gas turbines [23], photovoltaic arrays [24], converter transformer[25], and industrial ovens [26]. An important distinction has to be made between online and offline OSD. Offline detection awaits a complete time series as input and performs the operational state detection only after finishing the data collection process of a time series. Online detection, in turn, aims to periodically predict the operational state of a machine while still collecting new data. Past research focuses on online OSD. Consequently, OSD is most often seen as a simple time series classification instead of a time series segmentation problem. Various works about online OSD have been published; the proposed methods include deep learning [25, 27], supervised classification [24], and unsupervised clustering [24]. The use case presented in this paper, however, belongs to the field of offline OSD, as the data is solely analyzed after finishing the data collection process of the whole testing cycle of a pump. Offline OSD is only sparely found in literature. One example would be [23], which defines a method for detecting the number of different operational states in a gas turbine. To the best of our knowledge, no previous work covered a use case where OSD is interpreted as a time series segmentation problem and deep learning used for the segment labeling process. We call this type of task Operational State Segmentation (OSS) in the following chapters.

## 3   MODEL ARCHITECTURE

### 3.1  Problem Formalization

A general supervised multivariate time series segmentation problem can be formalized as follows: Let an evenly-spaced univariate time series with T timesteps be denoted by $X \in \mathbb{R}^T = [x_1, x_2, \ldots, x_T]$. Then a sensor-based multivariate time series can be formalized as two-dimensional array $X^S = [X_1, X_2, \ldots, X_S] = [[x_1^1, x_2^1, \ldots, x_T^1], \ldots, [x_1^S, x_2^S, \ldots, x_T^S]]$ where S denotes the number of sensors. In dense labeling, each timestep t comes with a distinct class label y, which has to be encoded to become applicable for deep learning. The encoded label $y_t^C$ for each time step t represents the label probabilities where the correct label has probability 1 and all other labels have probability 0. C denotes the number of different labels which can be assigned to a timestep. In the use case presented in this paper, a label represents the operational state of a hydraulic pump within a testing cycle. The total sequence of labels over all timesteps is given by $Y^C \in \mathbb{R}^{T \times C}$. The final output of the network, namely the label predictions, are denoted as $\hat{Y}^C \in \mathbb{R}^{T \times C}$. The task consists now of finding a function (model) f: $X^S \to \hat{Y}^C$ which minimizes a loss function $\ell(\hat{Y}^C, Y^C)$, representing the distance between the predicted and the true labels.

### 3.2  PrecTime Architecture Overview and Modules

The backbone of the PrecTime deep learning architecture is depicted in Figure 4. PrecTime is an end-to-end sequence-to-sequence approach, meaning that a whole time series can be labeled in one connected process. It builds upon the foundation of a two-module sliding window framework for sleep staging [4], but extends it by adding a third module for prediction refinement to enable precise segmentation. In turn, the PrecTime network consists of three modules, one for feature extraction, one for context detection, and one for prediction refinement. Furthermore, a time windowing splitting step at the start and a time-window merging step at the end have to be performed. Each of the three network modules serves a different purpose and plays an important part in the prediction process (confirmed in an ablation study in Section 5). The network belongs to the multi-objective network category, meaning it provides two different sorts of predictions, an intermediate one after the context detection module and a final one after the prediction refinement module. The intermediate predictions are included to ensure a stable training and, therefore, are only relevant for the training stage of the network. For evaluating the PrecTime performance and the inference, only the final prediction is relevant.

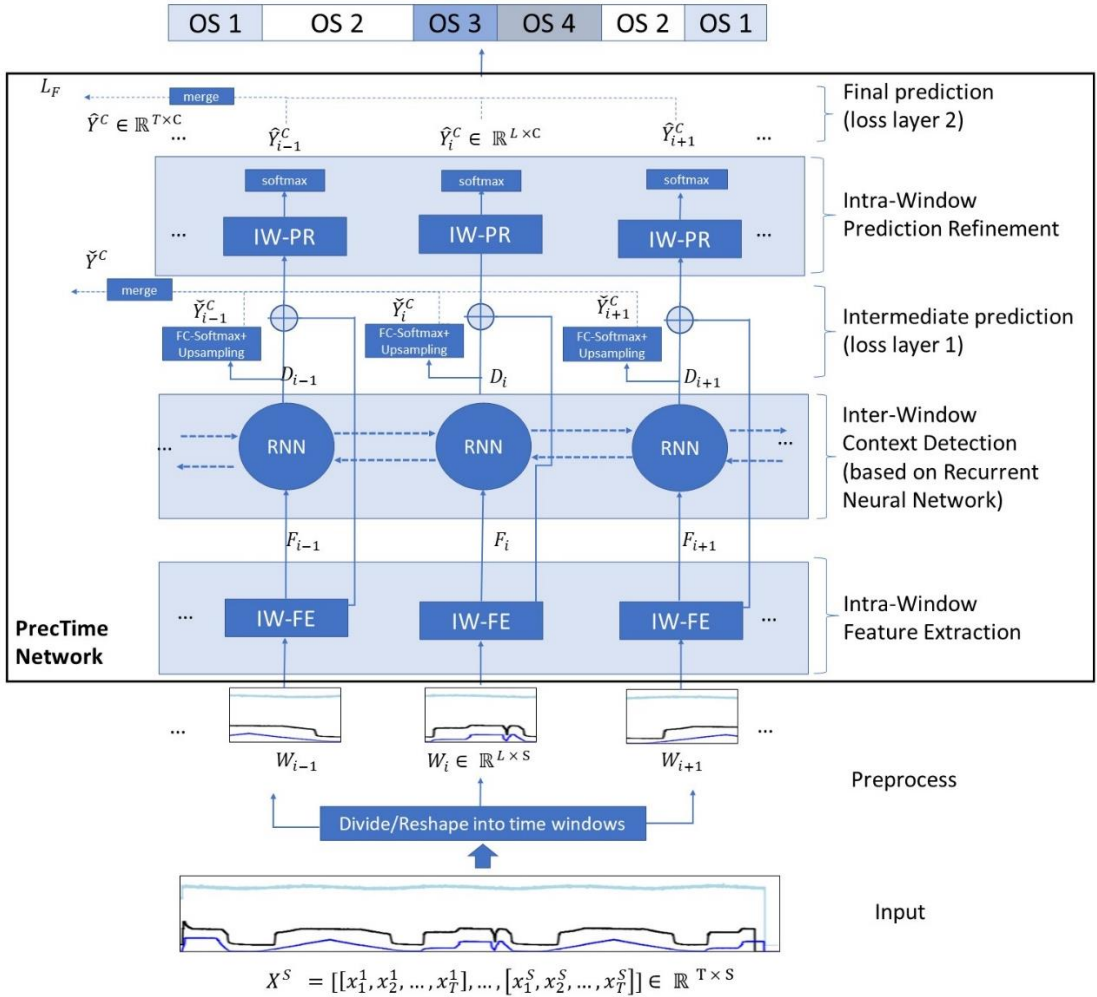

Fig. 4. Backbone of PrecTime network architecture

### Preprocessing

Before sending the data to the network, a time series $X^S$ needs be split up into N equally sized, non-overlapping time windows of length L. Accordingly, $X^S$ is transformed into a series of windows $W_i^S$, with i = 1,...,N where $W_i^S = [[x_{i1}^1, x_{i2}^1, ..., x_{iL}^1], ..., [x_{i1}^S, x_{i2}^S, ..., x_{iL}^S]] \in \mathbb{R}^{L \times S}$

### Feature Extraction Stage

The first module deals with classic feature extraction and is applied to each time-window in isolation. Specifically, it serves to find meaningful information within a time window and to compress this information to enable more accurate predictions at later stages. Finally, the feature array has to be flattened such that the output of the module is a 1D-feature vector $F_i$ for each Time Window $W_i^S$ with i = 1,...,N. In general, the concrete layer architecture of the module could be varied. However, previous research has shown that CNN-based architectures with additional pooling and drop-off layers performed well. In this paper, the architecture displayed in Figure 5 (left part), which was inspired by the SeqSleep+ network [18], is used. Through two parallel streams that highly differ concerning the dilation rate of the kernel, information of two different time scales can be captured. The pooling layer makes the information more compact while the drop-off layers are included to prevent overfitting.

### Context Detection Stage

The context detection stage consists of inter-window recurrent layers. It takes the feature vectors $F_i$ of all time windows i= 1,...,N as input and processes them within the context of to the information found in other time windows (context detection). The distinct feature vectors $F_i$ represent the different time states fed to the Recurrent Neural Network (RNN), enabling the network to detect valuable mid- and long-term dependencies that cannot be extracted when processing each time-window in isolation. Contextual information is especially useful in use cases where the time series segments have a certain order or high long-term temporal interdependence. The exact layer architecture is flexible; in this paper, a two-layer bidirectional LSTM is chosen to ensure that enough temporal context information can be captured, both in the forward- and backward pass. As the number of states the LSTM-layer has to process is limited to the number of time windows N, the vanishing gradient problem is very unlikely to pose a problem for the training process. Note that this distinguishes PrecTime from other LSTM-segmentation approaches where often each of thousands of data points is fed as distinct state into the LSTM creating a high risk for the vanishing gradient problem. The output of the module consists of a vector $D_i$ made up by the predictions of the different LSTM channels at each time step i.

### Prediction Refinement Stage

The final module enables the network to make one prediction per time point instead of only giving one label per time window. This module serves as connector between the time window-based architecture of the first two modules and the dense labeling output of the last layer. Like the feature extraction module, it processes each time step in isolation and therefore, classifies as intra-window module. The input at each timestep i constitutes a combination of the unflattened output of the feature extraction layer $U_i$ and the upsampled output of the context detection stage $D_i$. This way the module can include both the fine-grained spatial and temporal information within a time window and the inter-window contextual information. We propose the use of a CNN-based upsampling layer, which roughly mirrors a simplified version of the feature extraction module at the start (see Figure 5 right part). The output consists of a series of timestep-wise label probability predictions $\hat{Y}_i^C$ for each time window from 1 to N with $\hat{Y}_i^C = [[\hat{y}_{i1}^1, \hat{y}_{i2}^1, ..., \hat{y}_{iL}^1], ..., [\hat{y}_{i1}^C, \hat{y}_{i2}^C, ..., \hat{y}_{iL}^C]]$. To transform these dense label predictions in the right form, the last layer has to be fully connected (FC) with the number of nodes being equal to the number of different labels C. The applied activation function after the FC-layer is softmax. Afterwards, the time-window distributed output of the network is merged and transformed into the desired form $\hat{Y}^C = [[\hat{y}_1^1, \hat{y}_2^1, ..., \hat{y}_T^1], ..., [\hat{y}_1^C, \hat{y}_2^C, ..., \hat{y}_T^C]]$. Note that this allows for the comparison with the encoded ground-truth label sequence $Y^C$.

### Dual-Loss Architecture and Loss Function

The network is constructed as a dual-loss architecture, which highly affects the training process. In addition to the final label predictions $\hat{Y}^C$ of the network, an intermediate prediction $\tilde{Y}^C$ is also considered when calculating the loss and performing backpropagation. The intermediate prediction $\tilde{Y}^C$ is achieved by connecting the output of the second module $D_i$ to a fully connected layer with the number of nodes being equal C. Next, a softmax activation layer is used to get a label prediction for each window. Finally, the label prediction is upsampled by a factor equal to the window size L and the predictions of all windows are aligned sequentially to get $\tilde{Y}^C$. The total loss $\ell(\{\tilde{Y}^C, \hat{Y}^C\}, Y^C)$ then is captured by a function of both prediction layers $\{\tilde{Y}^C, \hat{Y}^C\}$ and the real labels $Y^C$ and calculated by a weighted sum method. The weight $w_F$ of final layer loss $\ell_F$ should be weighted higher than the weight $w_I$ of the intermediate loss weight $\ell_I$. This is done to boost and stabilize the weight updating of the first two modules and ensure that the base idea the network, namely that the accurate, but coarse prediction after the context detection module is further refined in the last module, is found in the training process. In turn, also the accuracy of the final predictions is expected to be increased.

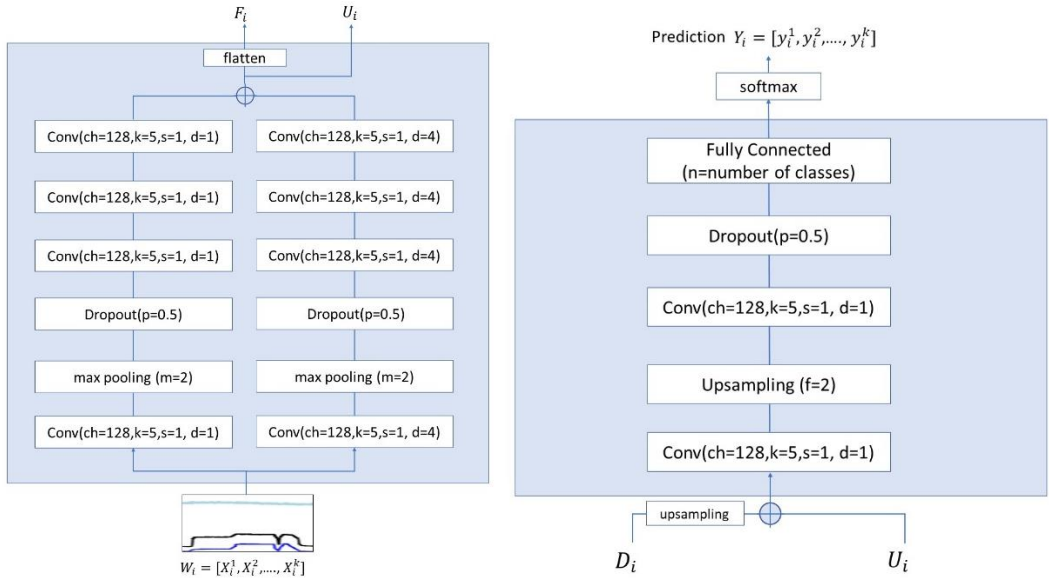

Fig. 5. Detailed Architecture of Feature Extraction Module and Prediction Refinement Module:
Convolutional Layers: Conv(ch = channels, k = number of kernels, s = stride ,d = dilation rate), Dropout Layer: Dropout(p = dropout probability),
Max Pooling Layer: max pooling(m = filter size), Upsampling Layer: Upsampling(f = upsampling factor)

## Network Parameters

When looking at the number of network parameters to be optimized, Table 1 lists the parameters broken down to the different layers. In total, there are almost 12 million trainable parameters (mainly caused by the first bidirectional LSTM-layer), which is in line with already existing networks for complex classification tasks that range from around 1-100 million trainable parameters.

TABLE I
NUMBER OF TRAINABLE PARAMETERS PER LAYER

| Module | Layer | Number of layers | Trainable parameter per layer |
|---|---|---|---|
| Feature Extraction | Conv(ch=128,k=5,s=1, d=1) | 2 | 5.888 |
| Feature Extraction | Drop-Out | 1 | 0 |
| Feature Extraction | Max-pooling | 1 | 0 |
| Feature Extraction | Conv(ch=128,k=5,s=1, d=1) | 6 | 82.048 |
| Context Detection | LSTM(100) | 2 | 5.200.400 |
| Context Detection | LSTM(200) | 2 | 320.800 |
| Loss Layer 1 | Fully Connected | 1 | 8.442 |
| Prediction Refinement | Conv(ch=128,k=5,s=1, d=1) | 1 | 210.048 |
| Prediction Refinement | Upsampling | 1 | 0 |
| Prediction Refinement | Conv(ch=128,k=5,s=1, d=1) | 1 | 82.048 |
| Prediction Refinement | Drop-Out | 1 | 0 |
| Loss Layer 2 | Fully Connected | 1 | 5.418 |
| **Total Network** | | | **11.852.420** |

## 4 DATASET AND EXPERIMENTAL SET-UP

PrecTime performance is evaluated on a novel hydraulic testing dataset. The technical background of the dataset and the experimental design is presented in the following section.

### 4.1 End-of-Line Testing in a Hydraulic Pump Factory

End-of-line testing constitutes a process in which the function of a product is tested to ensure its quality before its delivery. It is particularly common in the mobility sector, as product component failures can have enormous consequences in the field. End-of-line testing usually consists of measuring different values at predefined operational states and comparing them to a defined target state. If the measurements do not show any worrying signs or indicate any abnormality, the product is deemed to be functional and sent out. This article discusses an EoL testing use case in a hydraulic pump plant. Before beginning the testing cycle, a pump is mounted to a hydraulic testbench and nine different sensors are attached to the mounted pump, measuring hydraulic and mechanical characteristics over the course of the testing cycle. Visualizations of a hydraulic axial piston pump and the related hydraulic testbench for EoL testing are illustrated in Figure 6. The hydraulic pumps can tolerate a pressure level up to 500bar and are used in the field for travel drives. Technical details about the used sensors and the testing process cannot be disclosed due to confidentiality reasons.

(A)                                    (B)

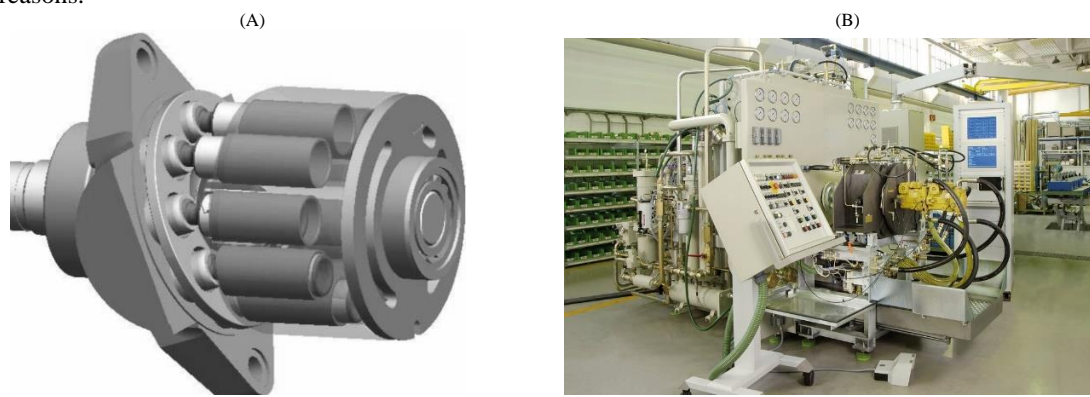

Fig. 6. Section A: Schema of hydraulic axial piston pump [28],
Section B: Hydraulic Testbench used for End-of-Line Testing (Source: Bosch Rexroth AG)

After the pump is mounted, a predefined testing cycle is started, which, in theory, is identical for all pumps of the same type. The testing cycle usually lasts, depending on how flawless the procedure works, between 3 and 7 minutes. Causes of the varying time range include unwanted repetitions or prolongations of test phases. The testing cycle consists of forcing the pump sequentially into different operational states. While it must be ensured that each specified operational state is reached at least once, deviations from the recommended order and the repetition of certain operational states (due to measurement problems) are very common.

After completing the testing cycle, the different operational states have to be evaluated individually to detect all potential faulty pumps. However, testbenches usually cannot flag a high number of different operational states and, therefore, the collected data of the entire testing cycle of a pump are only available as one long multivariate time series without or with only very few state labels. The measured values of the nine different sensors represent the different time-dependent variables.

At this stage, an accurate segmentation of the time series is necessary to automatically evaluate the various operational states. Inaccurate segmentation might lead to wrongfully assuming that state-specific tolerances are exceeded, characteristic curves look anomalous, and state-specific frequency calculations are biased. The segmentation is aggravated by the fact that the tested pumps are bi-directional, meaning that each operational state has to be evaluated in both flow-directions with the sensors not being able to distinguish between the two directions. The detection of the flow-direction of a specific operational state is therefore not possible by just focusing on measured sensor value information, but only by investigating the contextual information. For each pump, it is defined in which order the two flow-directions of one operational state appear within a test cycle. Therefore, depending on whether the operational state appears the first time or the second time within the time-series, the flow-direction of an operational state can be correctly assigned. This concept is illustrated in Figure 7 and poses an additional challenge for time series segmentation models. The use-case represents a typical offline segment labeling task which had never been investigated before in the operational state detection context.

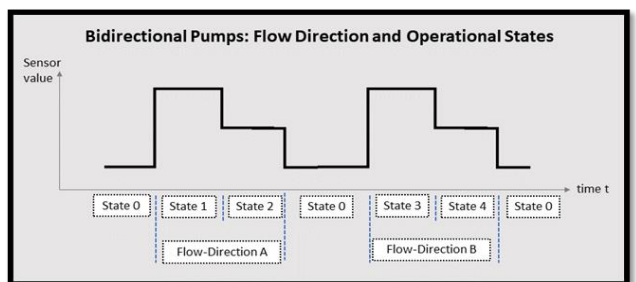

Fig. 7. Bidirectional Operating States:
The operational states 1/3 and 2/4 have identical sensor values. They can only be distinguished by order of their occurrence. Here, the testing cycle defines that the A-direction is tested first.

### 4.2 Hydraulic Pump End-of-Line Dataset

The Hydraulic Pump End-of-Line Dataset is made publicly available at "https://github.com/boschresearch/Hydraulic-EoL-Testing/". In this paper, we focus on the direct control (DC) pump subset. The subset consists of a total of 120 labeled samples; each sample is a multivariate time series representing the entire End-of-Line testing cycle of a unique pump. The operational state labeling was performed by a human operator. As the labeling process requires a significant amount of time, it is planned to be automated through the approach of this paper. The 120 samples are equally distributed to three pump versions (V35, V36, V38), with each pump version having 40 samples. There are faulty and good pumps in the dataset to confirm that the segmentation also works if anomalies in the measured values are found. The samples differ in length, with the majority of them lasting around 4-6 minutes. Figure 8 gives an overview of the composition of the data subset used in this article.

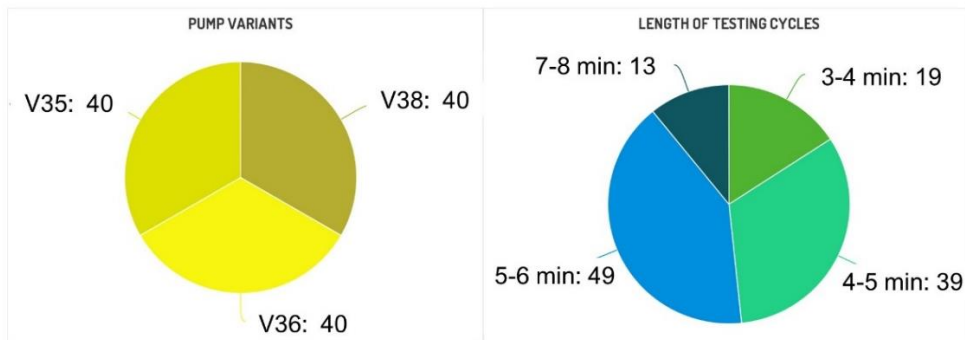

Fig. 8. Composition of HydraulicPump-EoL Dataset: cycle lengths and quality assessment

One sample includes nine different time series measured by nine different sensors. All sensors have a sampling rate of 100 Hz. Within a cycle, a maximum of 44 different operational states is distinguished. These states can be predefined functional test points, transition states between test points, characteristic curves, or special states (beginning, end, empty). These states already distinguish the flow direction, meaning that the same test point for two different flow directions is counted as two states. The states are integer-encoded with the integers ranging from one to 51. While usually a cycle includes all these states at least once, it can occur that certain transition stages can be skipped in some cycles or testing errors occur. Therefore, there is an expectation, but not a guarantee, to find all operational states within a cycle. The testing cycle for the different pump variants is very similar: for all variants, there is a maximum of 44 different states per cycle, and the operational states labels have for each variant an equivalent meaning. However, the detailed definitions of the operational states vary for the different pump variants. Consequently, in this paper, a separate model is trained and evaluated for each of the three variants. For confidentiality reasons, the sensor values are normalized (z-score standardization), and the used hydraulic, electric, and mechanical sensors are not further specified. Furthermore, no additional information about the background of the testing cycle is given to not violate corporate data protection issues. Figure 9 exemplarily shows a sample multivariate time series of one pump including the integer-encoded operational state labels. The state labeling was accomplished manually by an expert familiar with the testing process.

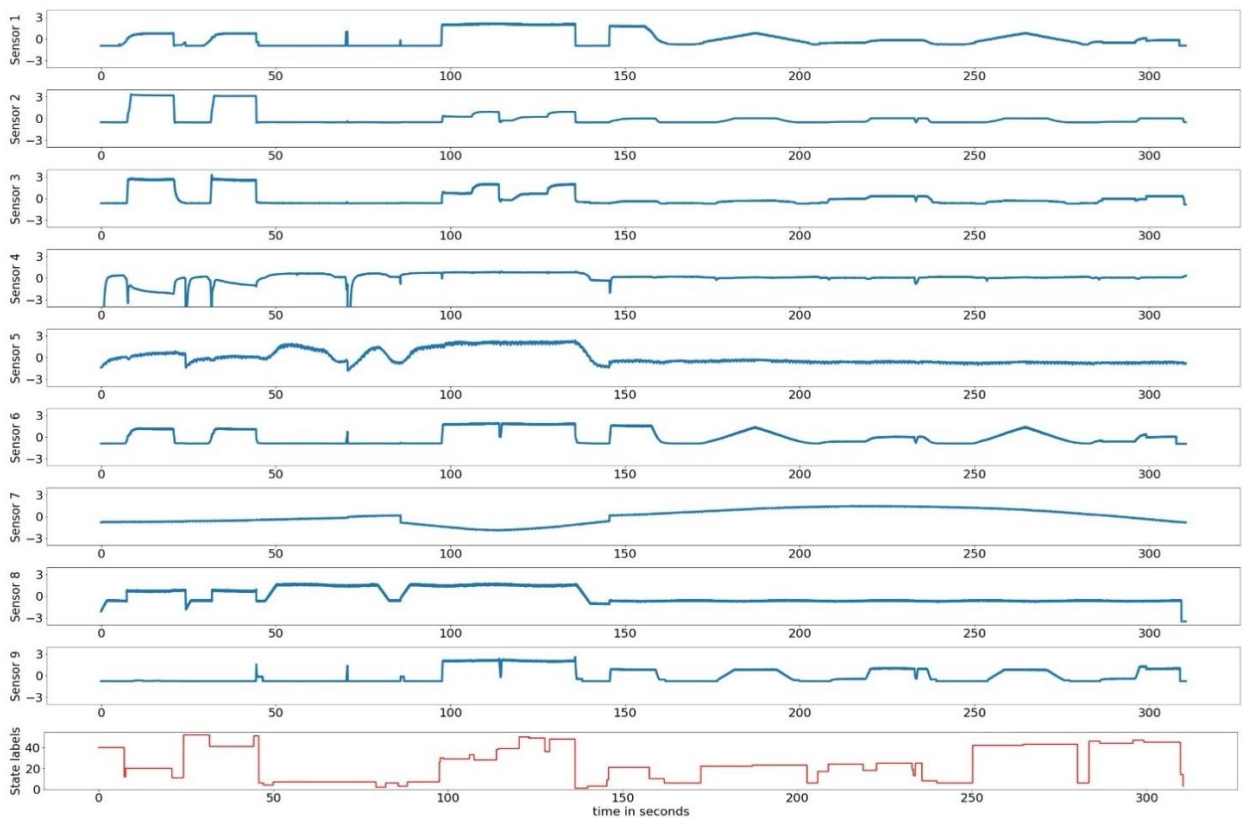

Fig. 9. Plotting one sample of the dataset including groundtruth state labeling (Sensors 1 to 9 are normalized)

*4.3 Experimental Set-Up*

## Data Preparation and Preprocessing

The experiment is designed to compare the segmentation performance of PrecTime to various baselines. To get robust results, the performance comparison is done for all three pump variants in the dataset. The 40 samples which exist for each variant are split into training (60%, 24 samples), validation (20%, 8 samples), and testing data (20%, 8 samples). The training data is used to train the network, the validation data is used for early stopping and comparing hyperparameter configurations, and the testing data is used to caluclate the final performance metrics. While the amount of samples seem less for a deep learning use case, it has to be considered that one sample usually includes the information of all 44 different operational states. In turn, the high information richness of a sample decreases the need for excessive number of samples.

Before the data can be used for deep learning, they have to be preprocessed. There needs to be an equal length for all samples. The testing cycle length varies from 3 to 7 minutes. Therefore, smallest value padding was used to create a uniform length of 450 seconds for each cycle. The data points added through padding were equal to the minimum value of each respective sensor, as the minimum values after normalization are representing zero in the original sensor measurement scales. For PrecTime and other sliding-window approaches, it is additionally necessary to split the testing cycles into equally-sized time windows. In this case, the length of a window was decided to be one second (100 values per window) and to use a non-overlapping time-windowing approach. This way it is ensured that the feature extraction module has enough information to extract valuable features and that the LSTM-based context detection module does not suffer from the vanishing gradient problem too heavily, as the LSTM has only to deal with 450 different time states.

## Metrics

In total we use four performance metrics to evaluate and compare the operational state labeling performance across the different tested networks. The first one is accuracy (1), a classis ML-metric, which simply states what percentage of the time stamps were labeled correctly. However, as accuracy can be misleading in case of highly-imbalanced class-labels, the Macro-F1-Score is additionally measured. The Macro-F1-Score (2) is calculated by simply taking the average of all class-specific F1-scores.

$$accuracy = \frac{number\ of\ correctly\ predicted\ data\ points}{number\ of\ data\ points} \qquad (1)$$

$$Macro\ F1 = \frac{1}{|C|}\sum_{c \in C} F1_c \ \ with\ F1_c = \frac{TP_C}{TP_C + \frac{1}{2}(FP_C + FN_C)} \qquad (2)$$

While these two metrics assess the general labeling performance, two additional metrics are introduced to focus on the exactness of the changepoint predictions, meaning timesteps, in which one operational state ends and another one begins. A predicted change point is assessed to be correct when it lies in the close proximity of an equivalent changepoint in the ground truth dataset. Both the ground truth changepoint and the predicted changepoint can only be considered as equivalent if the state transition takes place between the same operational states in both changepoints. Close Proximity is defined to be 20 timesteps, meaning the ground-truth equivalent changepoint is not allowed to be further away than 0.2 seconds from the predicted changepoint. This small tolerance is chosen due to the fact that a precise segmentation of operational states is necessary in our use case. Close Proximity may be defined larger in other domains. We define a metric called change-point-precision ($CP_{Prec}$), which measures the percentage of predicted change points being within the tolerated proximity of an equivalent ground truth change point (3). We complement this metric with the so-called changepoint-recall ($CP_{Rec}$) that represents the percentage of groundtruth changepoints that were correctly detected by the model in the sense that a predicted equivalent changepoint is in close proximity of the ground-truth changepoint (4). Like the names suggest, both metrics build on the idea of the typical binary classification metrics precision and recall and adjust them.

$$CP_{Prec} = \frac{number\ of\ predicted\ cp\ in\ proximity\ of\ equivalent\ groundtruth\ cp}{number\ of\ predicted\ changepoint} \qquad (3)$$

$$CP_{Rec} = \frac{number\ of\ groundtruth\ cp\ in\ proximity\ of\ equivalent\ predicted\ cp}{number\ of\ groundtruth\ changepoints} \qquad (4)$$

The four metrics allow assessing both the general labeling performance and the exactness of the predicted operational state transitions.

**Hyperparamtertuning**

The exact parameter settings were determined by a hyperparameter-tuning process performed before the final model training. To be able to have a fair comparison, the same parameter specifications were used for all three pump variants. For the hyperparameter tuning, a Random Search approach was chosen evaluating in total 25 different parameter combinations. The details and the final parameter specification can be found in Table II.

TABLE II
HYPERPARAMETER SEARCH AND BEST DETECTED COMBINATION

| Hyperparamter | Value Range | Best detected combination |
|---|---|---|
| Channels CNN | (64 - 320) | 256 |
| Channels LSTM | (100 - 200) | 200 |
| Kernel CNN | (3 - 10) | 5 |
| Final Layer Weight $W_F$ | (1 - 8) | 2 |
| Time-Window | (100-300) | 100 |

**Implementation and Training Process**

The model training and evaluation was performed on a NVIDIA Tesla K80 with 6 cores, 56 GB RAM and 380 GB disk storage running on a Linux operating system. The models were implemented in Python using the Tensorflow and Keras libraries. The training took place on Microsoft Azure ML instances to parallelize the training process and to increase the available computational power. Cross-entropy was chosen as the loss-function to be optimized for all trained networks. The used optimizer was Adam with a learning rate of 0.001. For each network, the training consisted of a maximum of 200 epochs. However, as some networks have shown a worsening performance after a specific threshold was reached, an early-stopping mechanism was implemented which stopped the training when there was no accuracy improvement in the validation dataset for more than 10 epochs. Since the number of training samples is very low, all training samples were used in each epoch to train the network (batch size = 24). After the network training, the final metrics (accuracy, F1-score, CP-precision, and CP-recall) were calculated based on the model performance on the test dataset.

**Implemented Baselines**

In total, we implemented five distinct baseline architectures which represent the range of the current state of the art in different domains. They were chosen in a way that a variety of deep learning approaches for time series segmentation were covered. For each architecture, the parameters were set like recommended in the respective papers, with the exception that the window length and stride of the time-window-based approaches were equally set to 100 for the sake of comparison.

- One-to-One LSTMS-CNN [13]: This classic and comparatively simple HAR-architecture is a time-window based approach. Each time window is classified in isolation and without context based on a LSTM-CNN architecture (one prediction per time-window)
- SeqSleep+ [18]: A network successfully applied to sleep staging using a many-to-many time window approach with a CNN-based architecture as intra-window feature extractor and a bidirectional LSTM-layer for the intra-window dependency detection.
- U-Time [2]: Dense labeling fully-convolutional network architecture which was first applied to sleep staging. It is the results of transferring the idea of the U-Net to the time series domain.
- Filternet [22]: Complex dense labeling network based on a CNN-LSTM architecture and successfully applied to an HAR-dataset.
- MS-TCN ++ [21]: A multi-stage (more than one loss-function), fully-convolutional architecture for dense labeling from the HAR domain, which expands the kernel-size exponentially in each layer. The input layer was changed to expect a 1D-input instead of a 2D input.

## 5   RESULTS AND DISCUSSION

### 5.1  Comparison of PrecTime and Baseline Performance

A summary of the experiment results can be found in Table III. PrecTime outperforms all baseline architectures, especially when considering the changepoint detection metrics. It performs excellently for all three pump variants in terms of correct labeling (accuracy, F1-score), as well as in the precise prediction of time segment changepoints (CP-precision and CP-recall). With an average accuracy of 96.1%, it exceeds the accuracy of the best baseline model (SeqSleep+ with 93.3%) by almost 3%. The difference becomes even clearer when comparing the changepoint-metrics in which PrecTime outperforms the state of the art with CP-precision of 59.5% and CP-recall of 75.6%, compared to the best-scoring baseline model MS-TCN++ with 36.7% CP-precision and 46.5% CP-recall. Since the testing process and operational state definitions of the three pumps are similar, the measured metrics only differ slightly for the three variants. The results show that, of the investigated models, PrecTime is best suited to realize precise operational state detection with stable performance across different product variants.

TABLE III

PERFORMANCE OF PRECTIME (ACCURACY, MACRO F1-SCORE, CP-PRECISION, CP-RECALL) COMPARED TO FIVE BASELINE-NETWORKS ACROSS THREE DIFFERENT PUMP VARIANTS

| Model | | 1-1 LSTM-CNN | U-Time | MS-TCN++ | Filter Net | Seq Sleep+ | PrecTime |
|---|---|---|---|---|---|---|---|
| Pump V35 | Acc | 0.547 | 0.595 | 0.918 | 0.459 | 0.933 | **0.958** |
| | F1 | 0.378 | 0.401 | 0.842 | 0.272 | 0.802 | **0.917** |
| | CP-Prec | 0.069 | 0.032 | 0.334 | 0.002 | 0.326 | **0.522** |
| | CP-Rec | 0.111 | 0.043 | 0.398 | 0.004 | 0.291 | **0.662** |
| Pump V36 | Acc | 0.602 | 0.785 | 0.929 | 0.464 | 0.939 | **0.966** |
| | F1 | 0.399 | 0.581 | 0.861 | 0.243 | 0.788 | **0.922** |
| | CP-Prec | 0.063 | 0.121 | 0.373 | 0.012 | 0.302 | **0.652** |
| | CP-Rec | 0.092 | 0.114 | 0.468 | 0.017 | 0.264 | **0.824** |
| Pump V38 | Acc | 0.577 | 0.598 | 0.932 | 0.399 | 0.926 | **0.959** |
| | F1 | 0.381 | 0.388 | 0.859 | 0.196 | 0.791 | **0.906** |
| | CP-Prec | 0.021 | 0.054 | 0.413 | 0.002 | 0.361 | **0.611** |
| | CP-Rec | 0.158 | 0.045 | 0.528 | 0.002 | 0.317 | **0.783** |
| Average of 3 pumps | Acc | 0.575 | 0.659 | 0.926 | 0.441 | 0.933 | **0.961** |
| | F1 | 0.386 | 0.456 | 0.854 | 0.237 | 0.794 | **0.915** |
| | CP-Prec | 0.051 | 0.069 | 0.373 | 0.005 | 0.330 | **0.595** |
| | CP-Rec | 0.120 | 0.067 | 0.465 | 0.008 | 0.291 | **0.756** |

The different modules of PrecTime work very efficiently together to both extract and consider both the spatial and temporal information found in the data, creating precise predictions. Figure 10 exemplarily shows the predicted labels of PrecTime for two pumps compared with the underlying ground truth labels. For comparison purposes, the predictions of two exemplary baseline networks (SeqSleep+ and U-Time) are also displayed. We can observe how well PrecTime performs the operational state segmentation task despite the high number of different states. There are less visible misclassifications compared to the baseline networks; most operational state labels were predicted equivalent to the ground truth.

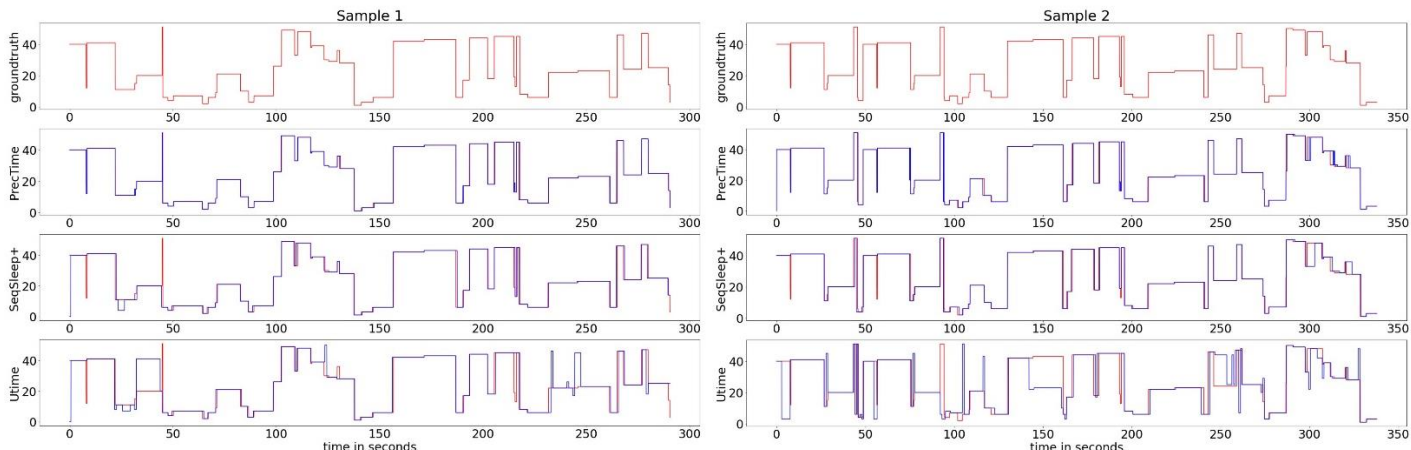

Fig. 10. Comparison of network label predictions (blue) and underlying groundtruth labels (red) for two sample pumps

Additionally, Figure 11 shows the evolution of the validation loss and the validation accuracy of the two loss layers. Interestingly, the intermediate loss layer performs better in the early stages of the training process, before being surpassed by the final loss layer in the later stages. This can be easily explained by the fact that the final loss layer is the result of three modules and thus, more network parameters need to be trained correctly compared to the intermediate loss until the label prediction becomes accurate. The difference between the intermediate layer accuracy and the final accuracy can be interpreted as the added value of the prediction refinement module.

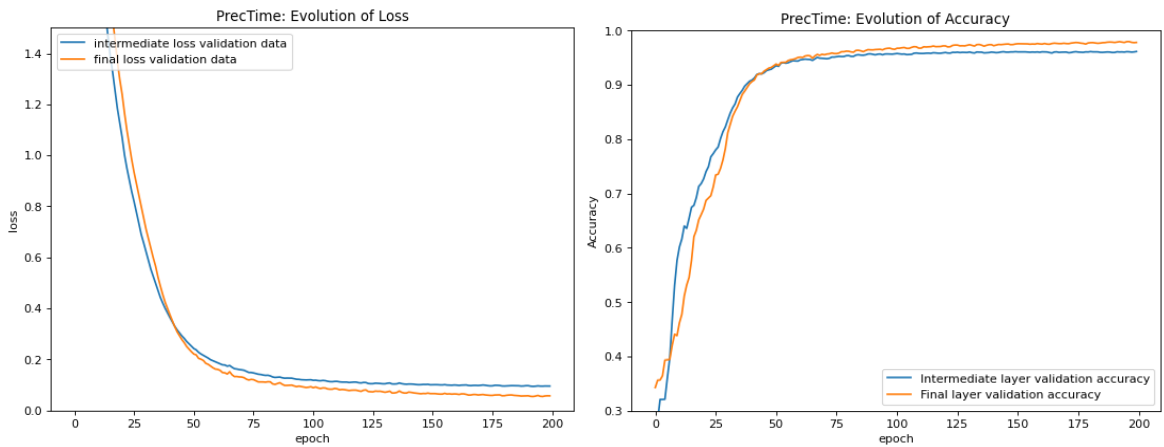

Fig. 11.  Evolution of key metrics during training process of PrecTime over the 200 epochs: validation loss and validation accuracy

When looking at the baselines, only MS-TCN++ and SeqSleep+ were able to create satisfying results close to PrecTime, whereas the other networks did not perform well at all for this use case. To explain the highly varying performance of the different networks, we have to look more deeply at the characteristics of the baseline architectures:

- One-to-one CNN-LSTM comes with one big problem: Since each time-window is classified without any context, long- and mid-term temporal information in the data cannot be captured. Therefore, certain operational states are confused and it is not possible to distinguish the flow direction of the states.
- U-Time is a very computationally intensive network and has a slow training progress. Additionally, the lack of an LSTM-module makes long-term context detection difficult.
- FilterNet combines CNN and LSTM layers, but in a different way compared to PrecTime. The LSTM-layer is not time window-based, but creates a prediction for each output timestamp. Therefore, the layer has to deal with 45,000 time steps as input, which creates the vanishing gradient problem. Even after 200 epochs, no real learning was seen in our use case.
- MS-TCN++ performs well, since in some convolutional layers kernels with very high dilation rate are used. They can partially balance out the lack of a RNN-module when it comes to long-term context detection.
- SeqSleep+ also performs well. However, the limited granularity of the prediction worsens the accuracy and particularly decreases the changepoint metrics.

None of the tested state-of-the-art networks was able to solve the operational state detection task as well as the PrecTime network. When looking at the runtime of the training process (Table IV) for 200 epochs, PrecTime showed a medium runtime of around 2h50min compared to the other networks. While especially complex dense labeling architectures like MS-TCN++ and FilterNet require long training times, the time-window based approaches have confirmed one of their major advantages, namely that they only need shorter training. Altogether, the main benefit of PrecTime lies in the clearly improved precision in detecting label changepoints and short transition phases, while the general labeling accuracy could only increase slightly compared to SeqSleep+ and MS-TCN++. However, the increased changepoint precision is highly significant in domains characterized by short state durations and a high frequency of label changes, such as EoL testing, as imprecise segmentation can quickly worsen the performance of follow-up algorithms like anomaly detection.

TABLE IV
MODEL TRAINING RUNTIME OF BASELINE NETWORKS AND PRECTIME

| Model-Runtime | Pump 1 | Pump 2 | Pump 3 |
|---|---|---|---|
| 1-1 LSTM-CNN | 1h59m | 1h35m | 1h32m |
| SeqSleep+ | 2h 26m | 2h 26m | 2h 18m |
| U-Time | 2h 5m | 2h 6m | 2h 5m |
| MS-TCN++ | 3h 39m | 3h 40m | 3h 42m |
| FilterNet | 12h 22m | 12h 59m | 12h 30m |
| PrecTime | 2h 48m | 2h 55m | 2h 51m |

### 5.2  Ablation Study

To evaluate and confirm the added value of each module in the PrecTime architecture, we performed an ablation study in which the full network was compared to reduced variations of the network. In each variant, one of the three modules was missing. The results are displayed in Table V. We can see that each module improves the performance of the network. In our industrial operational state segmentation use case, however, especially the context detection module is of high importance due to its ability to distinguish the flow directions of operational states.

- PrecTime_A1: The filter extraction module was left out, and instead, for each time window, just a flattening of the input was performed.
- PrecTime_A2: The context detection LSTM-module was left out and replaced by a fully-connected layer instead.
- PrecTime_A3: The prediction refinement stage was ignored; only the output of the intermediate output layer was regarded as relevant. The loss function, which was optimized during training, consisted only of the intermediate loss layer.

TABLE V
ABLATION STUDY OF PRECTIME

| Accuracy | Pump V35 | Pump V36 | Pump V38 |
|---|---|---|---|
| PrecTime (Original) | **0.958** | **0.966** | **0.959** |
| PrecTime_A1: no filter extraction | 0.939 | 0.944 | 0.929 |
| PrecTime_A2: no context detection | 0.74 | 0.76 | 0.71 |
| PrecTime_A3: no prediction refinement | 0.932 | 0.939 | 0.926 |

### 5.3 Application to Dataset from another Domain

To provide evidence for the claim that PrecTime is a general-purpose architecture, we applied it to a dataset from a different domain. We chose the popular HAR opportunity dataset, which contains labeled motion tracking sensor data [29]. The labels represent the activity of the human wearing the tracking sensors. We kept the network architectures and hyperparameter settings used in the hydraulic pump experiments. We only used the four available drill-recordings for our experiment (subjects 1 and 2 as training data, subject 3 as validation data, and subject 4 as testing data). From all available sensors, 113 channels were selected, as recommended in other works [29]. Missing values were interpolated. The annotations indicating the object moved by the right hand were selected as labels for the experiment. We had a setting with 113 input channels and 11 different labels. During preprocessing, zero-padding was used to get all instances to the length of 60000. The results are displayed in Table VI and show that, even in a fundamentally different setting and domain, PrecTime shows excellent results and outperforms the reference architectures.

TABLE VI
*SEGMENTATION PERFORMANCE OF ARCHITECTURES FOR OPPORTUNITY DATASET (DRILL RUNS)*

| Model | | 1-1 LSTM-CNN | U-Time | MS-TCN++ | Filter Net | Seq Sleep+ | PrecTime |
|---|---|---|---|---|---|---|---|
| Opportunity (Drill) | Acc | 0.617 | 0.487 | 0.612 | 0.794 | 0.605 | **0.915** |
| | F1 | 0.398 | 0.283 | 0.403 | 0.668 | 0.421 | **0.856** |
| | CP-Prec | 0.329 | 0.321 | 0.367 | 0.461 | 0.342 | **0.576** |
| | CP-Rec | 0.090 | 0.040 | 0.121 | 0.409 | 0.089 | **0.605** |

### 5.4 Further Research Directions and Limitations

The following section describes possible directions for future work and mentions some limitations of the paper which still have to be overcome.

#### Transfer Learning to reduce labeling effort

The manual labeling of operational states is very time-consuming and requires an experienced operator. Thus, it is of high importance that the ground truth labels are correct and precise, otherwise the learning of the network is hindered. To reduce the labeling time as well as the risk of labeling mistakes, it is desirable to reduce the number of labeled samples required for the training process. As the testing processes and operational states are similar across the pumps evaluated in this paper, the use of intra-domain transfer learning strategies can be evaluated. It can be checked whether a network, pre-trained with a high number of samples of one pump variant, performs well when being fine-tuned with a very small number of samples of another pump. If this is confirmed, it can be leveraged to enable efficient operational state detection in a high-variance factory.

#### Test PrecTime on other datasets from different domains

The PrecTime architecture performed well on the presented End-of-Line testing use case; however, it remains to be investigated how it compares to other baseline architectures when evaluating it on other datasets from other domains. PrecTime is generally applicable to all kinds of time series data with multi-phased character, and potential applications in other domains include manufacturing equipment monitoring, HAR, and speaker diarization. It further has to be checked whether the hyperparameter settings can be kept and transferred to other use cases, or whether a completely different hyperparameter configuration must be used. A detailed analysis should be conducted on how different parameters, such as window size, loss function, and training sample size, affect the performance of the model and on whether the results can be generalized to provide generally valid recommendations.

**Reduce weight of the model while keeping performance**

The network is a heavy-weight model with over 9 million trainable parameters. Further research could demonstrate whether and where the weight of the network can be reduced without significantly compromising its performance. This would be particularly desirable in high-variance production environments in which a large number of different models must be trained, as it would reduce training time.

**Use segmented operational states for follow-up algorithms**

The results of the operational state segmentation do not generate a direct benefit for a plant; however, they serve as an important base for preprocessing for further operations. Interesting follow-up tasks could include the automated detection of faulty pumps or the identification and elimination of redundant testing phases.

## 6 CONCLUSION

This paper proposed a novel sequence-to-sequence architecture for precise time series segmentation called PrecTime. The network is based on a classic time-windowing framework, but has been extended by a prediction refinement module and a dual-loss concept that was created by adding an intermediate output layer. Therefore, the network is enabled to make dense segment labeling predictions whose accuracy was successfully demonstrated when applied to a newly published industrial dataset. The use case investigated in this thesis dealt with the segmentation of operational states in the EoL testing process of hydraulic pumps, where the necessity of precise time series segmentation is high due to the crucial and sensitive follow-up tasks. The data used were collected in a real-world setting on hydraulic test benches with a high number of available sensors. We could show that, compared to other baseline architectures, the performance metrics of PrecTime significantly increased over all three tested pump variants; very high prediction accuracies of around 96% could be reached compared to the selected state-of-the-art networks, which achieved accuracies ranging from 44% to 93%. The added value of all three modules of the architecture was successfully evaluated via an ablation study, which showed that, especially, the LSTM-based context-detection module is of high importance in this use case. Future work should investigate how PrecTime performs on other datasets and check if transfer learning strategies could be used to minimize manual labeling effort in high-variance production environments.